\documentclass[journal]{IEEEtran}
\usepackage{graphicx}
\usepackage{booktabs}
\newcommand{\ra}[1]{\renewcommand{\arraystretch}{#1}}

\ifCLASSINFOpdf
\else
\fi
\begin{document}
%
\title{Privacy-Preserving Object Detection \& Localization Using Distributed Machine Learning: A Case Study of Infant Eyeblink Conditioning}

%
%
%

\author{
        Stefan~Zwaard, 
        Henk-Jan Boele,
        Hani Alers,
        Christos Strydis, 
        Casey Lew-Williams,
        and~Zaid Al-Ars}
\thanks{S. Zwaard is with The Hague University of Applied Sciences, The Hague, The Netherlands, e-mail: S.L.W.Zwaard@hhs.nl}
\thanks{H.-J. Boele is with the Princeton Neuroscience Institute, Princeton, NJ 08544, USA, e-mail: hboele@princeton.edu}
\thanks{H. Alers is with The Hague University of Applied Sciences, The Hague, The Netherlands, e-mail: Hal@hhs.nl}
\thanks{C. Strydis is with the Erasmus University Medical Center, Rotterdam, The Netherlands, e-mail: c.strydis@erasmusmc.nl}
\thanks{C. Lew-Williams is with the Department of Psychology at Princeton University, Princeton, NJ 08544, USA, e-mail: caseylw@princeton.edu}
\thanks{Z. Al-Ars is with the Delft University of Technology, Delft, The Netherlands, e-mail: z.al-ars@tudelft.nl}
\thanks{We thank Prof. S.S.-H. Wang (Princeton Neuroscience Institute) for intellectual and financial support.}

%
%

{Zwaard \MakeLowercase{\textit{et al.}}: Privacy-Preserving Object Detection \& Localization Using Distributed Machine Learning: A Case Study of Infant Eyeblink Conditioning}
%



\maketitle

\begin{abstract}
Distributed machine learning is becoming a popular model-training method due to privacy, computational scalability, and bandwidth capacities. 
In this work, we explore scalable distributed-training versions of two algorithms commonly used in object detection. A novel distributed training algorithm using Mean Weight Matrix Aggregation (MWMA) is proposed for Linear Support Vector Machine (L-SVM) object detection based in Histogram of Orientated Gradients (HOG).
In addition, a novel Weighted Bin Aggregation (WBA) algorithm is proposed for distributed training of Ensemble of Regression Trees (ERT) landmark localization. Both algorithms do not restrict the location of model aggregation and allow custom architectures for model distribution. For this work, a Pool-Based Local Training and Aggregation (PBLTA) architecture for both algorithms is explored. The application of both algorithms in the medical field is examined using a paradigm from the fields of psychology and neuroscience -- eyeblink conditioning with infants -- where models need to be trained on facial images while protecting participant privacy. Using distributed learning, models can be trained without sending image data to other nodes. The custom software has been made available for public use on GitHub: https://github.com/SLWZwaard/DMT. Results show that the aggregation of models for the HOG algorithm using MWMA not only preserves the accuracy of the model but also allows for distributed learning with an accuracy increase of 0.9\% compared with traditional learning. Furthermore, WBA allows for ERT model aggregation with an accuracy increase of 8\% when compared to single-node models.  

\end{abstract}

\begin{IEEEkeywords}
distributed machine learning, computer vision, facial images, landmark detection, privacy protection.
\end{IEEEkeywords}

%
\IEEEpeerreviewmaketitle

\section{Introduction}
%
%
%
%

\IEEEPARstart{D}{istributed} machine learning allows models to be partially trained on a local basis and then aggregated into a new model, for example on a central server~\cite{konen2015federated}. For research involving images or videos of human participants, this method serves the important purpose of protecting an individual's privacy by keeping the data local to the collection point~\cite{enthoven2020overview}. However, this approach has other advantages, including distribution of the required processing power needed for machine learning, which eliminates the need for heavy central servers, as well as reduction of data communication bandwidth requirements for transferring large amounts of data between nodes.

In order to enable distributed training, algorithms need to be developed for effective model aggregation that result in little (or no) loss in accuracy. In this paper, we focus on aggregation of a Linear Support Vector Machine (L-SVM) classifier-based object detection using a Histogram of Orientated Gradients (HOG) feature extractor, the HOG-algorithm~\cite{hog, felzenszwalb2010object}, as well as a landmark localization algorithm called Ensemble of Regression Trees (ERT)~\cite{kazemi2014one, Loh2011}. Both are common algorithms used in face- and feature-detection applications~\cite{zafeiriou2015survey, Bakker2017RealtimeFA}. The HOG-algorithm is used to detect objects such as human faces or road signs. For each object and its orientation, a specific model needs to be trained. For localizing landmarks on an object, such as dots around a person's eye, ERT is used to place a predefined shape of landmarks on the object. The predefined shape is then shifted and warped into place over multiple iterations. This ERT model needs to be trained for each specific situation or context within a dataset.

In order to train new models, annotated datasets are needed. However, these datasets should be specific for the conditions of the test setup to achieve the most accurate results. For face detection, specific datasets might be needed for multiple facial orientations or ages. For general face and object detection, publicly available datasets are often available. However, if specific images introduce noise or other variability, such as uncommon objects or equipment visible on the face, publicly available datasets do not help. Privacy is also a major limiting factor for image availability due to, for example, GDPR-compliance issues~\cite{chassang2017impact}.

In this work, new distributed-training algorithms for the HOG-algorithm and ERT are explored to allow training of new models across multiple nodes without sharing the original data. For the HOG-algorithm, we introduce a Mean Weight Matrix Aggregation (MWMA) training algorithm, while for ERT models, a Weighted Bin Aggregation (WBA) training algorithm is proposed. For both algorithms, a Pool-Based Local Training and Aggregation (PBLTA) distribution architecture is examined, where models can be shared with a pool to all participating parties. Without loss of generality, we prototype and test the two algorithms using a cutting-edge case which motivated this work in the first place and is borrowed from the fields of psychology and neuroscience, called eyeblink conditioning (EBC). EBC is a behavioral, Pavlovian-training~\cite{pawlow1927} experiment where the participant trains a response to a repeated stimulus, providing a general biomarker for neuro-developmental diseases~\cite{ReebSutherland2015EyeblinkCA}. Privacy concerns are currently limiting the training of new models in EBC, where we use computer vision to automate the eye-tracking experiments. These cutting-edge experiments are currently being conducted as a joint project between the Princeton Neuroscience Institute (PNI), the Department of Psychology at Princeton University, and the Department of Neuroscience at the Erasmus Medical Center (EMC). EBC measurements of both adult and infant populations are taken through the use of camera equipment, rather than physically attached sensors, in order to increase patient comfort. 

Previous work has shown that it is possible to detect the blinking of an eye in video footage in real time~\cite{Bakker2017RealtimeFA}. This process works by using a GPU-accelerated version of the HOG-algorithm-based face detector and a multithreaded ERT landmark localizer, which can then be used to calculate eyelid closure based on Eye Aspect Ratio (EAR)~\cite{soukupova2016real}. This process works with high accuracy on an adult population using the AFLW database~\cite{aflw2011}. However, to improve accuracy on an infant population in an EBC test setting, where an air puff device is visible on the participant's head, new models need to be trained on both face detection and landmark localization.

Due to privacy concerns related to the images of infants' faces, participating institutes cannot readily share the required data for training.  
Written consent is needed from infants' parents, which severely limits experimental-data availability. Moreover, new GDPR laws in the EU make sharing of data between the EU and the US even more challenging~\cite{phillips2018international}. However, our approach does not require multiple research teams to have the actual data; rather, just the final trained model inference results. Thus, distributed training of the needed models is introduced to overcome participant privacy concerns and to make collaboration possible in the first place.

This paper makes the following contributions:
\begin{itemize}
\item MWMA, a new scalable algorithm to allow distributed training of the HOG object-detection models using mean matrix averaging, independent of model-distribution architecture and without privacy concerns.  
\item WBA, a new scalable algorithm for the distribution of ERT model training using output averaging, independent of model-distribution architecture and without privacy concerns.
\item Exploration of a Pool-Based Local Training and Aggregation (PBLTA) architecture to share the locally trained sub-models or aggregated models between multiple research organizations.
\item Achieving an increase in accuracy when creating models using the distributed approach compared to traditional learning for the HOG-algorithm and compared to a single-node model for the ERT models.
\end{itemize}

This paper is organized as follows. Section~\ref{sec:background} discusses the background and significance of the approach we introduce in this research. Section~\ref{sec:related_work} describes related work in the use of distributed training for the ERT and HOG-algorithms. Section~\ref{sec:face_distribute} presents a method to distribute the training of object-detection models, using infant eyeblink conditioning as a test case, including measures of its accuracy. Section~\ref{sec:landmark_distribute} presents a method to distribute the training of landmark localization models and measures its accuracy. Then, Section~\ref{sec:ditribute_arch} presents a pool-based approach for practical model distribution.

\section{Background} \label{sec:background}

\subsection{Traditional centralized model training}
In traditional model training, all available training data are aggregated in a central location and used to train a target model, typically from scratch. The specific way a model is trained and the way the training data are formatted depends on the type of machine-learning algorithm used. For example, for industrial machine learning~\cite{diez2019data,helmi2019}, if data from multiple sensors or locations are used, all data must be sent to a central location for training, which generally requires high-bandwidth communication as well as a high-end infrastructure for storage and processing capacity~\cite{yao2018big, peteiro2013survey,  maji2018reduction}.

\begin{figure}
    \centering
    \includegraphics[width=\linewidth]{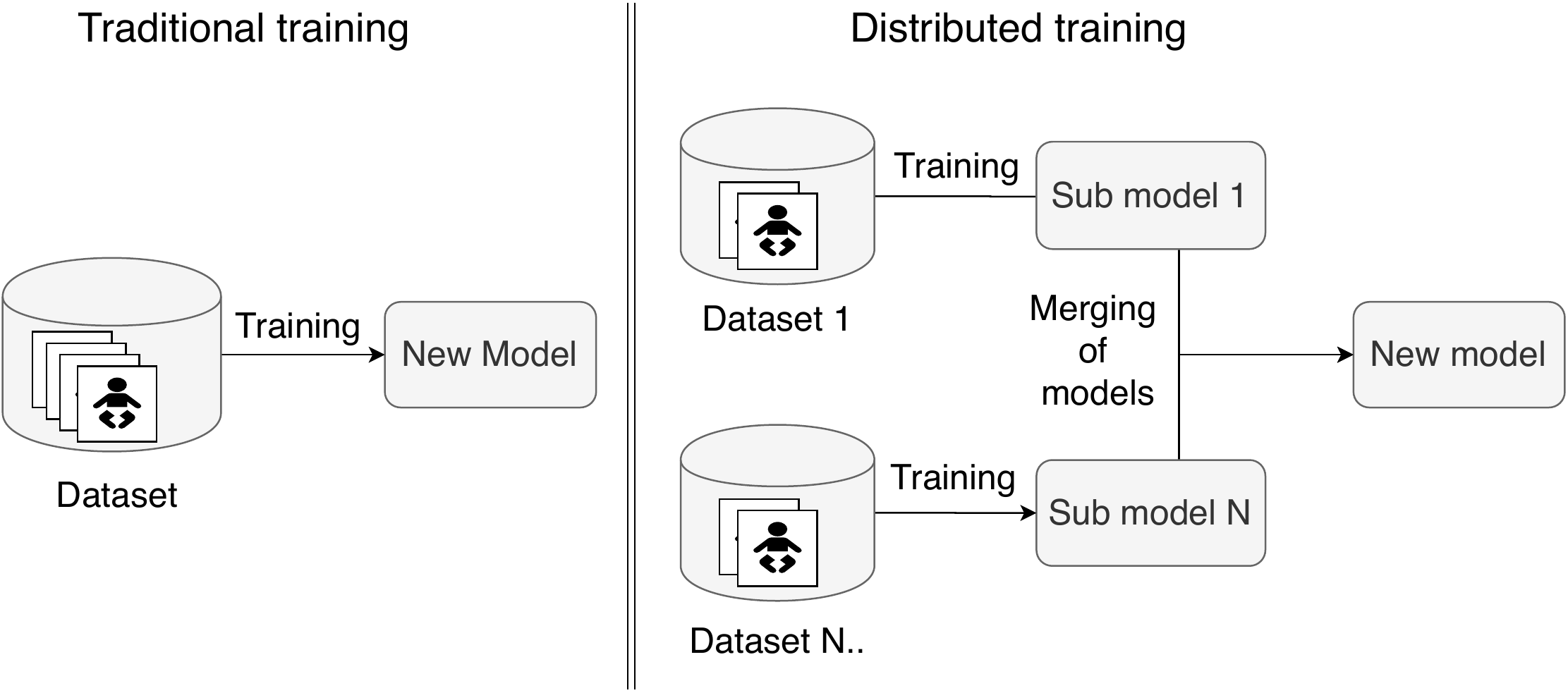}
    \caption{Difference between traditional model training (where a new model is trained from a single large dataset) versus distributed model training (where multiple partial models are first trained on locally available data before merging into a new, centralized model).}
    \label{fig:background}
\end{figure}

In this work, with EBC as a case study, the current version of the software used to process video data to detect eyeblinks requires two models: one for face detection using the HOG-algorithm, and one for landmark localization using ERT to position and adjust 68 landmarks on facial images.
Our baseline implementation is based on a centralized model training approach, where both models are trained on annotated facial data in a traditional way (see Figure~\ref{fig:background}). The training data are collected at both the EMC and PNI, requiring that facial data must be shared between the two teams. These facial data are not time-sensitive and do not require high-end infrastructure or high-speed bandwidth for sharing, but the data cannot be shared freely due to considerations about participant privacy. International data-sharing agreements with multiple institutions are required in order to enable meaningful communication between research teams.

\subsection{Distributed model training}
Distributed machine learning is gaining popularity in sectors ranging from industry to the medical domain. Example use cases can be found in industrial process control~\cite{yao2018big}, the oil and gas industry~\cite{kejela2014predictive},  cardiovascular-disease prediction~\cite{grebla2006distributed}, medical imaging~\cite{chang2018distributed} and the prediction of post-radiotherapy dyspnea~\cite{eurocat2017}. In the industrial sector for instance, traditional machine learning relies on heavy central-computer systems capable of storing and processing data received from multiple sensors~\cite{maji2018reduction}. The switch to distributed learning is then made to save cost and allow more use of local processing power~\cite{yao2018big, peteiro2013survey,  maji2018reduction}. For the medical sector, however, distributed learning is also interesting for its privacy-protection capabilities~\cite{beaulieu2018privacy, chang2018distributed, eurocat2017}. 

Distributed machine learning can be implemented in many ways~\cite{yang2019federated, kairouz2019advances}, depending on the type of machine-learning algorithm used and approach of distribution. Some forms of distributed learning depend on retraining existing models using locally available data, while others merge sub-models together to create a new model. Models can be merged centrally, as is the case in federated machine learning, or locally. In the present work, the approach of creating local sub-models and merging them into a new main model (see Figure~\ref{fig:background}), using a local model aggregation from a central pool storage, is explored for both SVM as well as ERT models. The central pool offers a hybrid solution where sub and main models are stored on a central server, but aggregation is done locally depending on selected sub-models. This approach is preferred as each participating node can select which sub-models to use. This allows each individual node to create a main model fitted to its specifications, and to prevent unwanted addition of poorly performing sub-models to bring down main model performance. In the EBC case study, this would allow models to be created for adults only, infants only, or a mix of adults and infants. Specific sub-models trained for differences in test setup or participants could also be chosen at will. 

\subsection{Eyeblink conditioning}
\begin{figure}
    \centering
    \includegraphics[width=\linewidth]{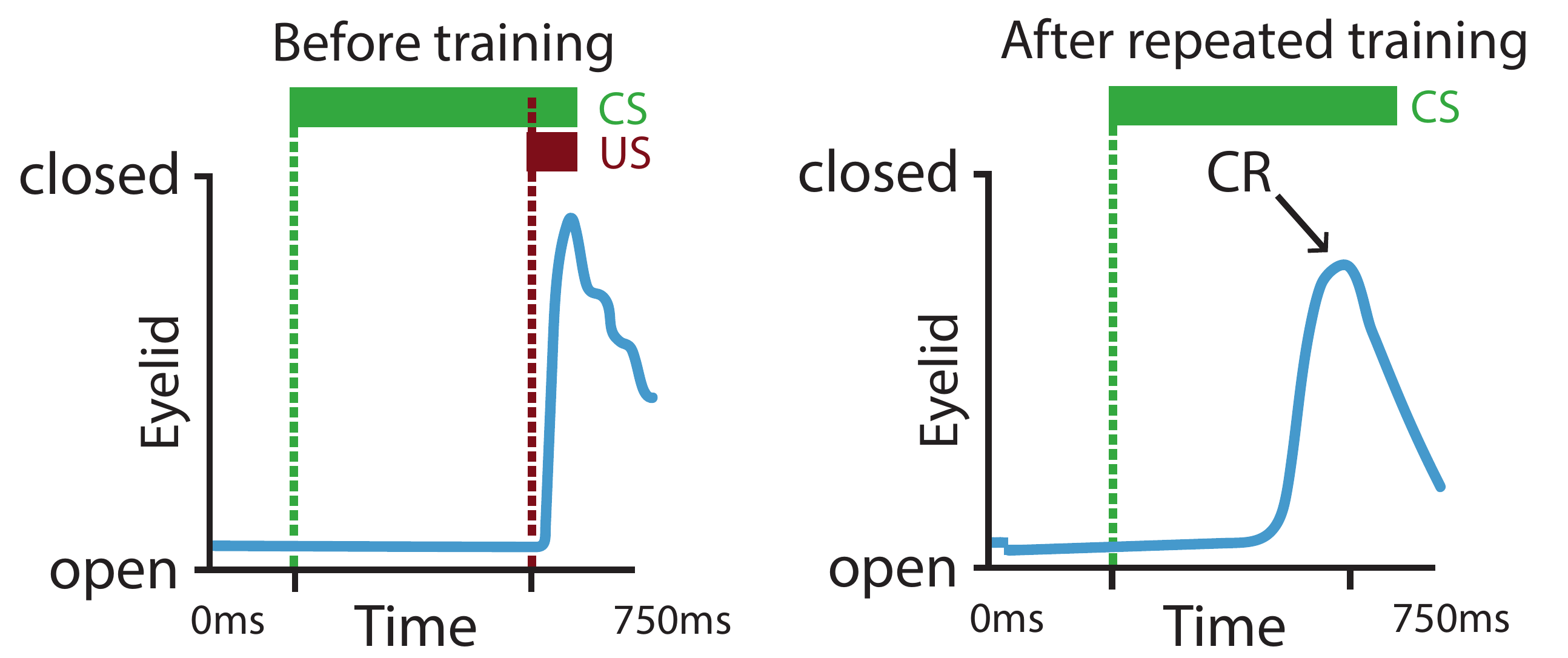}
    \caption{Conditioned (right) and unconditioned (left) responses and stimulus of an EBC test where a participant is trained to blink upon introduction of an auditory tone (CS) by pairing the tone with a puff of air in the eye (US).}
    \label{fig:ebc}
\end{figure}

Eyeblink conditioning (EBC) is a non-invasive behavioral test that is widely used in the fields of psychology and neuroscience~\cite{ReebSutherland2015EyeblinkCA}. EBC is based on Pavlovian conditioning~\cite{pawlow1927}. In the original work on Pavlovian conditioning, dogs learned the association between an auditory cue and the introduction of food. The introduction of food, called the stimulus (US), naturally lead to the production of saliva, which is called the unconditioned response (UR). Repeated pairing of the auditory cue, called the conditional stimulus (CS), with the US gradually led to the development of a conditioned response (CR), i.e., salivation. EBC is almost identical to this classic paradigm introduced by Pavlov but, in this case, the US consists of a mild air puff applied to the eye, which causes the eye to close in response (UR). Repeated pairing of the auditory cue (CS) and the air puff (US) leads to an eyelid closure, the CR, in response to the auditory cue, as shown in Figure~\ref{fig:ebc}. Typically, the onset of the auditory cue precedes the onset of the air puff by several hundred milliseconds. Standard outcome measures for EBC include the speed of acquiring the association, EBC kinetics (CR amplitude, CR timing), and the retention/extinction of the association. 

EBC testing provides several measures of how we learn in our daily lives, including but not limited to associative and motor-sensory learning. It is known now that the essential memory formation for EBC takes place in the cerebellum. Several lines of evidence suggest that EBC (in combination with other tests) can be used as a general biomarker for neuro-developmental diseases~\cite{ReebSutherland2015EyeblinkCA}. For example, autism spectrum disorder (ASD) is known to be linked to cellular dysfunction~\cite{autism2014}. Studies indicate that individuals with ASD show lower learning scores during delay testing in association-based tasks~\cite{autism_ebc2015}.
EBC testing is traditionally done with potentiometers~\cite{schreurs1997lateralization}, EMG~\cite{bracha1997patients}, magnetic distance measurement techniques~\cite{Koekkoek2002}, and low-level infrared or laser emitters~\cite{Marsh1979}. However, for infants and young children, many of these techniques are impractical, expensive, or require facially attached instruments~\cite{Anneli2014}. Therefore, in this work we use an image processing approach combined with landmark localization algorithms. The use of the object-detection and landmark-localization algorithm in EBC will be clarified in Section~\ref{sec:face_distribute} and~\ref{sec:landmark_distribute}. 

\section{Related Work} \label{sec:related_work}
\subsection{privacy-aware SVM object-detection training}
Overcoming privacy concerns when training SVM models for object detection has been addressed in prior work. In the work of Kitayama et al.~\cite{kitayama2019hog}, SVM models for face detection can be trained on encrypted images, without decryption of the images. This is done by using a changed HOG feature-extraction process that allows Encrypted then Compressed (EtC) images~\cite{watanabe2015encryption, kurihara2015encryption, kurihara2017encryption} as input. Extracted HOG features are comparable to those extracted from unencrypted images. This allows all data to be collected on a central location for traditional training of SVM models for object detection. However while the EtC encryption has currently no known vulnerability, it can be hacked with brute-force attacks~\cite{chuman2018security}. 
To allow image-processing work to be stored on a public Cloud, the work of Yang et al.~\cite{yang2017privacy} also allows extraction of HOG features of encrypted images, using Vector Homomorphic Encryption (VHE) on images. This achieved comparable results to using standard HOG features while the server calculating the features does not have the plain image. Similarly, the work of Wang et al.~\cite{wang2016sechog} uses a Somewhat Homomorphic Encryption (SHE) on images and has HOG feature-extraction methods for both single- and multi-server setups. Both the VHE and SHE are semantically secure as only negligible information can be gained from the chiper text~\cite{naehrig2011can} and require the recipient of the encrypted images to perform brute-force attacks to decipher the encrypted images~\cite{wang2016sechog}. In these works, a method was found to extract the HOG features without needing the source images in plain text. These features can then be used to train SVM models for object detection, or in our case study for EBC, a face detector. At the moment of this writing, there is no known way to reverse-engineer these HOG features back to the original images, thus forming no direct privacy risk to sharing annotated HOG features between organizations for SVM model training. However, there are methods that use HOG features for face recognition~\cite{deniz2011face} or to extract biometric information like gender from these features~\cite{castrillon2016using}, indicating that not all personal data are lost during feature extraction. Therefore, to prevent current and potential future privacy concerns, in the event that HOG features can be reverse-engineered, our work does not require sharing the HOG features between collaborating organizations. Only sub-models trained from multiple images are shared for the creation of a new model. Although this new model achieves higher accuracy when used on the original training images, indicating whether or not the image is part of the training set ~\cite{papernot2016towards}, it cannot lead to extraction of training data from the model itself.

\subsection{Privacy-aware ERT landmark localization training}
For the training of decision trees, data can be distributed across multiple nodes, for example because of privacy concerns for data sharing~\cite{du2002building} or if the total dataset is too large for practical central processing~\cite{ouyang2009induction}. Here, we use distributed training, which allows data to be used for training without sharing it with other nodes -- an idea that has been explored in existing literature. In the work of Du et al.~\cite{du2002building}, new decision trees are trained from multiple datasets by having each node calculate its own splits and using the best splits for the new tree. The best splits are calculated using the Scalar Product Protocol with the help of an untrusted third-party server. This server also does not have access to the final model. However, their method is limited to two nodes. Vaidya et al.~\cite{vaidya2008privacy}, using a privacy-preserving distributed ID3 algorithm, allow tree training across multiple nodes, while increasing the needed computation time and communication rounds. Ye et al.~\cite{ye2009stochastic} achieved distributed learning of Stochastic Gradient Boosted Decision Trees (GBDT) through the use of MapReduce and the Hadoop framework, reducing training time when data are distributed across multiple nodes. However, privacy of local data is not preserved as the central system has access to local training data. While these studies discussed the distributed training of trees from portioned data, distributed training of trees can also be accomplished through aggregation of locally trained trees into an ensemble. Multiple forms of tree-ensemble creation are discussed in the work of Ouyang~\cite{ouyang2009induction}. Ensembles can e.g.~be created either through boosting~\cite{lazarevic2002boosting}, bagging~\cite{chawla2003distributed} or random forests~\cite{breiman2001random}. Another solution is found by training trees of the new model locally, and then combining the trees into a singular new tree. This has been investigated using a Fourier-analysis-based technique~\cite{park2001fourier}, a Meta-learning framework in research by Prodromidis~\cite{prodromidis2000meta}, and Fisher's linear discriminant as discussed in the work of Ouyang~\cite{ouyang2009induction}.
For our investigation, we distribute the training process of the ERT model by having each local node train its own ensemble. These ensembles are then aggregated into a new ensemble, where the trees from each local node form their own group with a weighted bin for their output. These groups participate in the final output calculation through the use of an extra weighted layer, averaging the output of all bins. By using this model-creation method, each local ensemble can reach ideal accuracy during training on its own data model. No direct sharing with other nodes is needed. Models can be created from chosen nodes, for example based on their local performance, allowing dynamic formation of new ERT models. 

\section{Face-detection model distribution} \label{sec:face_distribute}

\subsection{Traditional face-detection algorithm}

The HOG-algorithm responsible for face detection and localization, was originally implemented using the DLib-Library~\cite{dlib09} and accelerated on GPU~\cite{Bakker2017RealtimeFA}. The HOG-algorithm from DLib is an implementation of the work of Dalal et al.~\cite{hog} with later inclusion of the downscaling pyramid features from the work of Felzenszwalb et al.~\cite{felzenszwalb2010object}. A HOG-based feature extrapolation from given images is used as input for a L-SVM-based sliding window classifier, to classify an area of 80x80 pixels as a face based on its trained models. The face detector from DLib uses 5 models trained on the Labeled Faces in the Wild (LFW) face dataset~\cite{huang2007labeled, Huang2014LabeledFI}. Each model represents its own facial orientation: frontal, faces turned left and right, and frontal with the head tilted left and right. Since a single model cannot be trained to include multiple facial orientations (these represent different objects for the L-SVM), inclusion of dedicated models for each facial orientation is needed. Furthermore, since human faces are right/left symmetric, a mirrored version of the left faces in the dataset was used to train the right faces model. 

\begin{figure}
    \centering
    \includegraphics{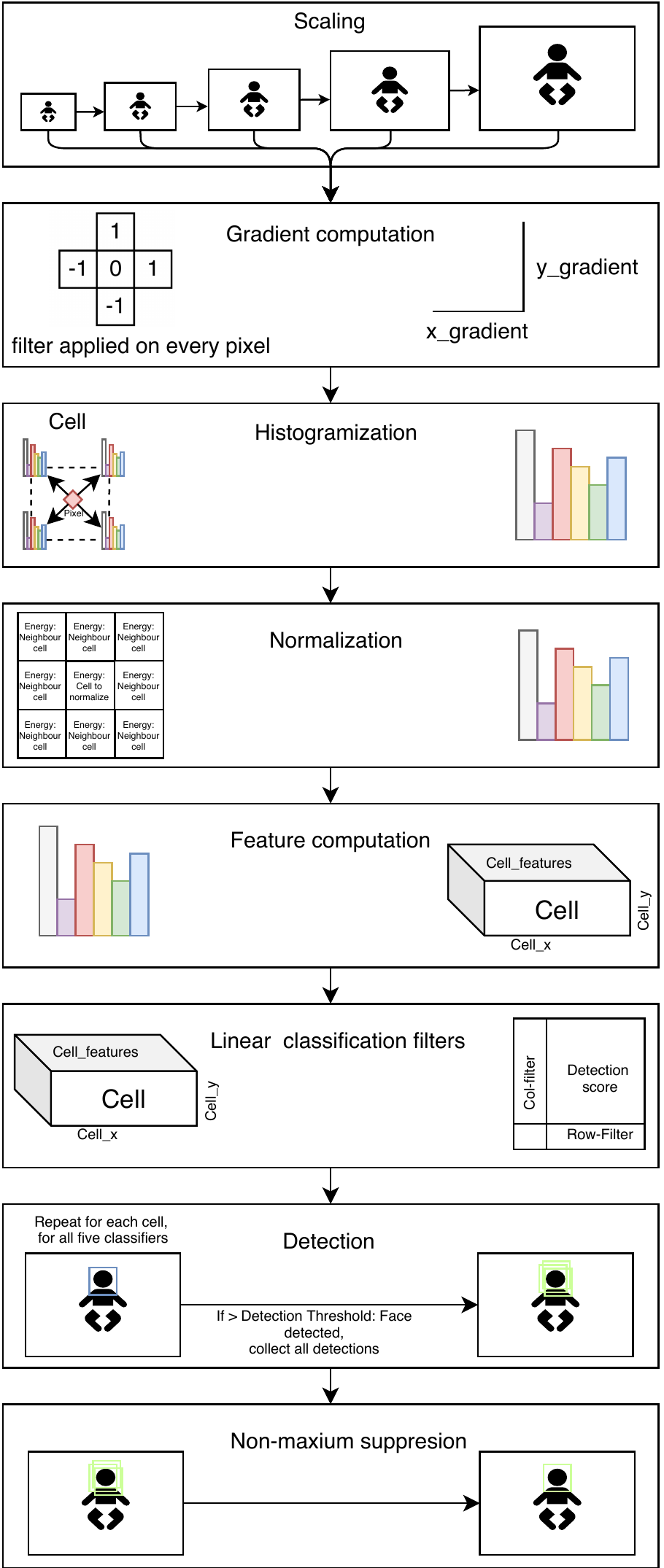}
    \caption{Face detection steps using the HOG-algorithm. HOG-features are extracted and used by a L-SVM sliding window classifier for face detection.}
    \label{fig:hog}
\end{figure}

The technical process of face detection, consisting of feature extraction and a detection process described in~\cite{felzenszwalb2010object}, can be summarized in the following 8 steps, which are also visualized in Figure~\ref{fig:hog}.

\begin{itemize}
\item \textbf{Scaling}: The original image is scaled down with factor 5/6 in iterations using bilinear interpolation until either the width or height of the image is smaller than those of the detection window (80$\times$80). Note that each following step is done on all iterations of the image.
\item \textbf{Gradient computation}: For every pixel in the image, the gradient is computed by applying the finite difference filter [-1,0,+1] for the x-direction at its transpose for the y-direction. The gradient orientation is computed as $\tan^{-1}(\frac{grad_y}{grad_x})$ and discretized into one of eighteen signed directions. The gradient magnitude is calculated as $\sqrt{grad^2_x+grad^2_y}$. \item \textbf{Histogramization}: The image is then divided into cells of 8$\times$8 pixels. Each cell is described by a histogram of eighteen bins, corresponding to the eighteen discrete gradient orientations of the pixels. Each pixel adds its gradient magnitude to the bins corresponding to this gradient orientation of four surrounding histograms. Bilinear interpolation is used to scale the contribution to each of the four histograms based on the location of the pixel.
\item \textbf{Normalization}: The energy value of each cell is based on the  eighteen bins, using the sum of nine pairs of opposite orientations squared. The energy of each cell is computed with $\sum^8_{n=0}(hist\_bin_n + hist\_bin_{n+9})^2$. The energy value is then normalized using the energy values itself and its eight surrounding cells.
\item \textbf{Feature computation}: The ﬁnal feature vector of each cell contains 31 features: 18 signed normalized histogram bins, 9 unsigned normalized histogram bins (where opposite orientations have been added together), and 4 gradient energy features, capturing the cumulative gradient energy of square blocks of surrounding cells. This completes the feature extraction phase: the original (scaled down) image is divided into cells of 8$\times$8 pixels, and each of those cells is described by a total of 31 features. We hereafter refer to this converted image as the feature image.
\item \textbf{Linear classification filter}: The classifier is trained with a face size of 80$\times$80 pixels, or 10$\times$10 cells, totaling 3100 features, which are used as input for the classifier. Each of these features is multiplied by a certain weight. This is done for every area of 10$\times$10 cells in the feature image. The multiplication of the weights is done in two steps. First, a row filter is applied to every feature of every cell, which multiplies the same feature of the ten horizontally neighboring cells with a different weight. The value of each feature of each cell is now the weighted accumulation of its ten horizontal neighbors. Next, the same process is carried out for every cell using a column filter for the ten vertically neighboring cells using the new accumulated values. When this is done, each cell contains a total of 31 features, each representing a weighted accumulation of the same feature of its 10$\times$10 neighboring cells. The total accumulation of these features is the detection score, and the image representing these detection scores for each cell is called the saliency image. As previously stated, there are five classifiers, for five different rotations of the face, each with its own filter values. Therefore, this whole process is repeated five times; once for every classifier.
\item \textbf{Detection}: The detection score of each cell is compared to the threshold of the classifier. If it is higher than the threshold, the cell is the center of an area of 10$\times$10 cells which is classified as a face. This is done for all five classifiers.
\item \textbf{Non-maximum suppression}: To prevent multiple detections of the same face, detections with an overlap of more than 50\% are eliminated, and the one with the highest detection score is kept.
\end{itemize}

\subsection{Distributed face-detection algorithm}
The HOG-algorithm requires a trained model, specifically one for each type of object and orientation. For the EBC case study, face-detection models need to be trained. When using the DLib implementation of the HOG-algorithm model training, a traditional approach, a new model is trained using a set of images and XML annotation of relevant or to-be-ignored areas. All data that need to be included for model training need to be available locally. As the training data of the EBC case study require sensitivity about issues of privacy, models need to be trained without sharing the training data. In this section, an average-based distributed training approach, the Mean Weight Matrix Aggregation (MWMA) algorithm, is proposed to enable training of a new model without sharing training data.

\begin{figure}
    \centering
    \includegraphics[width=\linewidth]{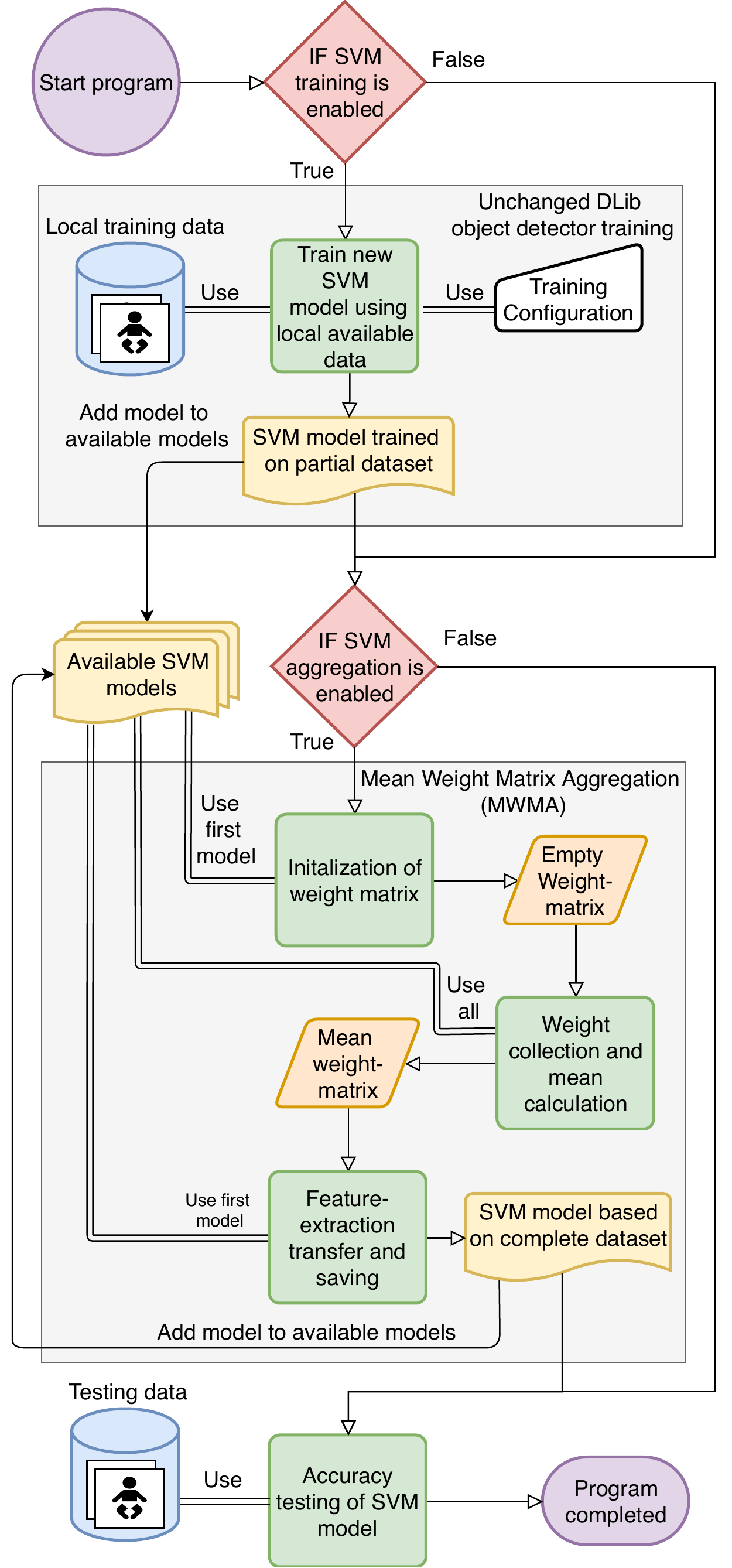}
    \caption{Object detection model training and Mean Weight Matrix Aggregation (MWMA) overview for the HOG-algorithm. A new model can be trained from a selection of local training data, adding the available models together with models trained on other nodes. Models are combined by the MWMA to create a new model, therefore making use of all available training data without sending data between nodes.}
    \label{fig:distrib_hog}
\end{figure}

The MWMA algorithm, shown in Figure~\ref{fig:distrib_hog}, has been developed based on the DLib implementation of model training for the HOG-algorithm. 
Models are trained using the default DLib process, using locally available training data. The MWMA algorithm is able to combine locally trained models from participating nodes into a new model, which represent the full training data used, thus increasing accuracy. Where the MWMA algorithm is performed depends on the model's distribution architecture, which is covered in Section~\ref{sec:ditribute_arch}. The performance of the MWMA algorithm is scalable, with a complexity of O(N) when models are trained across N nodes, as each node can train its model in parallel and aggregation time is negligible compared to training.

The process of model combination for the HOG-algorithm (using the MWMA algorithm) is based on a mean weight-matrix and feature-extractor from the used models. The steps taken for the creation of a mean weight-matrix and a new model are visualized in Figure~\ref{fig:distrib_hog} and described in the following steps.
\begin{itemize}
\item \textbf{Weight-matrix initialization}: Based on the first model read by the system, a new empty weight matrix is created. This matrix is used during the linear-classification filter step in the detection process. The weight matrix of the first model decides the size of the new matrix. The size of the matrix of a model depends on the detection size in the image. For our system, we use models trained for 80x80 pixels. Models with different detection sizes are incompatible. 
\item \textbf{Weight collection and mean calculation}: For each unit in the matrix, the weight value of the same unit for all models is collected. A mean is calculated by ${\displaystyle A={\frac {1}{n}}\sum _{i=1}^{n}p_{i}={\frac {p_{1}+p_{2}+\cdots +p_{n}}{n}}}$, where A represents the new aggregated value of the unit in the matrix, p represents the unit of each sub-model, and n represents the number of sub-models used in the aggregation process. This is carried out for the entire matrix. Time scales are computed linearly with the number of sub-models used for aggregation. 
\item \textbf{Feature extraction, transfer and saving}: The type of feature extraction used by the new model, in this case the HOG feature extraction system, and other related settings like the 80$\times$80 window size, are copied from the first sub-model. These and the new weight matrix are saved as a new model. The size of the new model does not increase with the number of sub-models used. 
\end{itemize}

In order to test the MWMA algorithm and the effectiveness of model combination by means of averaging, the accuracy of a created model by combining sub-models is measured and compared to a model trained traditionally on the same dataset. Accuracy is measured using recall and precision, where recall describes how many of the existing faces were indeed detected, while precision describes how many of the detected faces were indeed real faces and not false positives. Recall and precision are calculated as follows.

\begin{equation}
\mathit{recall}=\frac{\mathit{TP}}{\mathit{TP} + \mathit{FN}} \ \ \ \ \ \ \ \ \ \ 
\mathit{precision}=\frac{\mathit{TP}}{\mathit{TP} + \mathit{FP}}
\end{equation}

Here, TP stands for True Positive, which represents a correct detection made of a face. FP (False Positive) is incorrectly detecting an area as a face, such as a face detected on the dots of a participant's clothing or the air puff device. FN (False Negative) is a failure to detect an actual face correctly. 

\subsection{Evaluation results}

The algorithm is tested using the DLib annotated version of the LFW database~\cite{huang2007labeled, Huang2014LabeledFI}, containing 2323 images, the same dataset used to train the default DLib model. In order to simulate distributed nodes, the dataset is split into 7 segments. Six segments, P1 trough P6, are for training, where each has 333 images. The last segment, consisting of 325 images, forms the test dataset. Sub-models are trained using each of the 6 parts separately, and iterations of parts are combined, e.g., P1+P2 representing the training data of both parts being available for traditional training. The traditional training with all data is represented by Pall. Then models are created using the MWMA algorithm, by combining sub-models until all models are used, e.g., COM(P1+P2) representing a model made from combining P1 and P2.

For the HOG-algorithm training process, the training parameters from Table~\ref{tab:svm_config} were used. The sub-models are then combined in a different order to check possible influences of aggregation order dependency. The use of weighted-model combination is also tested by multiplying the presence of the same model. This allows a model from a selected dataset to influence the averaging process more than other models in order to weigh the end model to the selected dataset. All models are then tested on the test dataset to provide recall and precision results.

\begin{table}\centering
  \ra{1.3}
        \caption{Training settings used by for distributed SVM testing 
        } \label{tab:svm_config}
\begin{tabular}{@{}lclc@{}}
    \toprule 
     \textbf{Setting} & \textbf{Parameter}\\
      \midrule 
    C                   & 5.0 \\ 
    epsilon             & 0.01 \\ 
    target size         & 80$\times80$ \\ 
    upsamle amount      & 0 \\ 
    \end{tabular}
\end{table}

\begin{table}\centering
  \ra{1.3}
        \caption{Accuracy results for training and MWMA model combination for the hog-algorithm. 
        }
    \label{tab:hog_res}
\begin{tabular}{@{}lclc@{}}
    \toprule 
     \textbf{ Dataset} & \textbf{Recall} & \textbf{Precision}\\
      \midrule 
    LFW-DLib-P1               & 0.890  &   0.997   \\ 
    LFW-DLib-P2               & 0.874  &   1    \\ 
    LFW-DLib-P3               & 0.885  &   1  \\ 
    LFW-DLib-P4               & 0.885  &   1  \\ 
    LFW-DLib-P5               & 0.896  &   0.997  \\ 
    LFW-DLib-P6               & 0.901  &   1  \\ 
    LFW-DLib-P1+P2            & 0.885  &   1    \\ 
    LFW-DLib-P1+P2+P3         & 0.879  &   1        \\ 
    LFW-DLib-P1+P2+P3+P4      & 0.885  &   1   \\ 
    LFW-DLib-P1+P2+P3+P4+P5   & 0.894  &   1     \\ 
    LFW-DLib-Pall             & 0.877  & 1     \\ 
    LFW-DLib-COM(P1+P2)       & 0.888  & 0.997      \\ 
    LFW-DLib-COM(P1+P2+P3)    & 0.882  & 1      \\ 
    LFW-DLib-COM(P1+P2+P3+P4)    & 0.885  & 1      \\ 
    LFW-DLib-COM(P1+P2+P3+P4+P5)    & 0.885  & 1      \\ 
    LFW-DLib-COM(P1+P2+P3+P4+P5+P6)    & 0.885  & 1      \\ 
    LFW-DLib-COM(P5+P3+P1+P2+P6+P4)    & 0.885  & 1      \\ 
    LFW-DLib-COM(P2+P1+P6+P4+P3+P5)    & 0.885  & 1  \\ 
    LFW-DLib-COM(P1+P2+P3+P4+P5+P6+P6)    & 0.890  & 1      \\ 

      \bottomrule 
    \end{tabular}
\end{table}

Accuracy test results of the HOG-algorithm training and MWMA combination algorithm are stated in Table~\ref{tab:hog_res}. For most tested models, a precision of 1 is measured, which means no False Positives are generated during the testing process for that model. However, this does not indicate a high accuracy as the recall rates are significantly lower. The recall results differ for most models. The table shows that the model combined from all partial models achieves a higher recall rating of $\approx$0.885, around 0.9\% higher than the Pall model trained traditionally on the complete dataset. This indicates that the MWMA algorithm to create models using an averaging method is able to achieve results that are comparable to (if not better than) the traditional approach. One reason that MWMA is able to achieve higher accuracy might be the improved generalization capabilities of the newly created model compared to the models trained using the traditional approach. This improvement has been observed in distributed learning methods before, as in Konecny et al.~\cite{konen2015federated, konen2016federated} and others.

Lastly, as accuracy information does not differ between order variations of the combined models, the proposed system is also deemed order-independent. Therefore, there is no need to take the order of model combination as a consideration while deploying this method in the field. As a result, the distributed model deployment approach discussed in Section~\ref{sec:ditribute_arch} is order-invariant as well. In the final model combination, a bias is created towards the P6 model. The increase in accuracy of the resulting model could be accredited to the higher accuracy of the P6 model compared to the other partial models.

\section{Landmark localization model distribution} \label{sec:landmark_distribute}

\subsection{Traditional landmark localization algorithm}
The landmark localization process is used to identify the shape of specific landmarks on an object. In the EBC case study, it is used for facial landmarks, by tracing the locations of 68 landmark dots on the facial image and placing them around the jawline, mouth, nose, and eyes. The locations of these landmarks are then used for further analysis. For an EBC experiment, eyeblinks are detected by calculating an Eye Aspect Ratio (EAR)~\cite{Bakker2017RealtimeFA}, using the following formula~\cite{soukupova2016real}: EAR$=(||$LM44$-$LM48$||+||$LM45$-$LM47$||)/($2$\times||$LM43$-$LM46$||)$, where the LM\# represents the used landmark location. This gives an eyelid openness percentage for each processed frame of the video. This percentage is then normalized based on the entire EBC trial, where Normalized $=$ 100$-($EAR$-$ MIN\_EAR$)\times($100$/$MAX\_EAR$)$, resulting in eyelid closure percentages. When plotted against frame capture time, this results in an eyelid closure graph, e.g., the left plot in Figure~\ref{fig:ebc}. This can be used for blink detection, or other EBC parameters like CR amplitude.

\begin{figure}
    \centering
    \includegraphics[width=\linewidth]{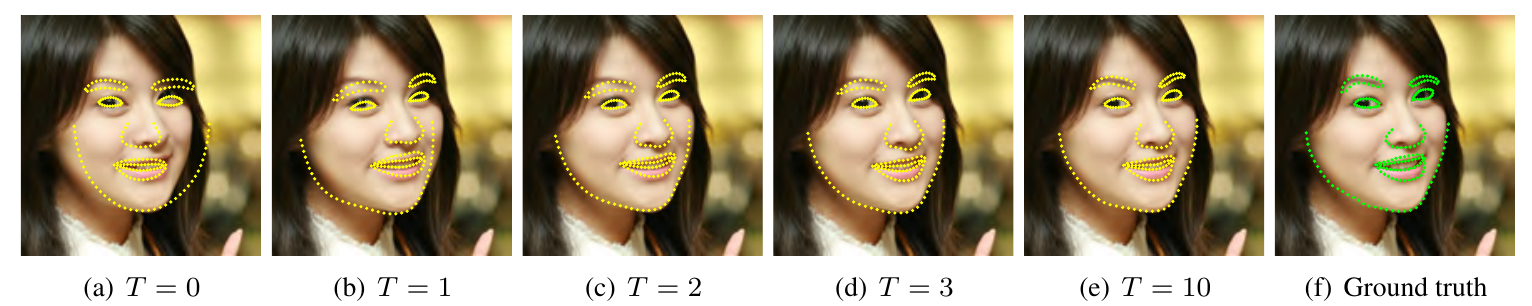}
    \caption{Overview of the iterative ERT process of adjusting the initial shape into the final facial form, where T represents the current iteration~\cite{kazemi2014one}}
    \label{fig:faces}
\end{figure}

Landmark localization uses the Ensemble of Regression Trees (ERT), a cascade-regression-shaped predictor implementation found in DLib~\cite{Bakker2017RealtimeFA}, which is in turn based on the work of Kazemi et al.~\cite{kazemi2014one}. The process of landmark localization starts by placing an initial shape estimate over the center of the faces detected by the face-detection algorithm. This shape estimate is represented by a specific configuration of the 68 landmarks calculated from the mean of all landmark configurations of a training set of facial images. Then a forest of regression trees gradually shifts the position of the landmarks toward the actual facial features using calculated features and pixel intensity. This is an iterative process, where feature calculation and landmark shifting using regression trees is done for each level of the cascade, as visualized in Figure~\ref{fig:faces}. The iterative process is summarized in the following steps~\cite{Bakker2017RealtimeFA}.

\begin{itemize}
\item \textbf{Initialization}: Initialize the landmark shape estimate. This is the pre-trained mean of all landmark shapes in the training set. 
\item \textbf{Feature computation}: Calculate the similarity transform between the estimation of the shape at the current cascade level and the original mean-shape estimation. The feature points are indexed relative to the initial mean shape and undergo the same transformation to calculate their position relative to the current shape estimate. Based on this transformation, new locations of the feature points are calculated for the regression trees.
\item \textbf{Regression-tree estimation}: Transverse each regression tree in the forest based on pixel intensity differences and threshold values, and add the results of their leaf values to the current landmark shape estimate.
\item \textbf{Repeat}: Repeat the feature computation and regression tree estimation step for each level of the cascade.
\end{itemize}

When using the default ERT model from DLib, also known as the shape predictor model, the forest consists of 500 regression trees, used by each of the 15 levels of the cascade. Each of these 500 trees from the forest has, by default, a depth of 5 layers leading to 16 possible outcomes, also known as leaves. In the layers leading up to the leaves, splits are used to make a decision, either to a next split or to the final leaf for that tree. The forest size, tree depth, and the number of levels in the cascade depend on the training configuration.

\subsection{Distributed landmark localization algorithm}

\begin{figure}
    \centering
    \includegraphics[width=\linewidth]{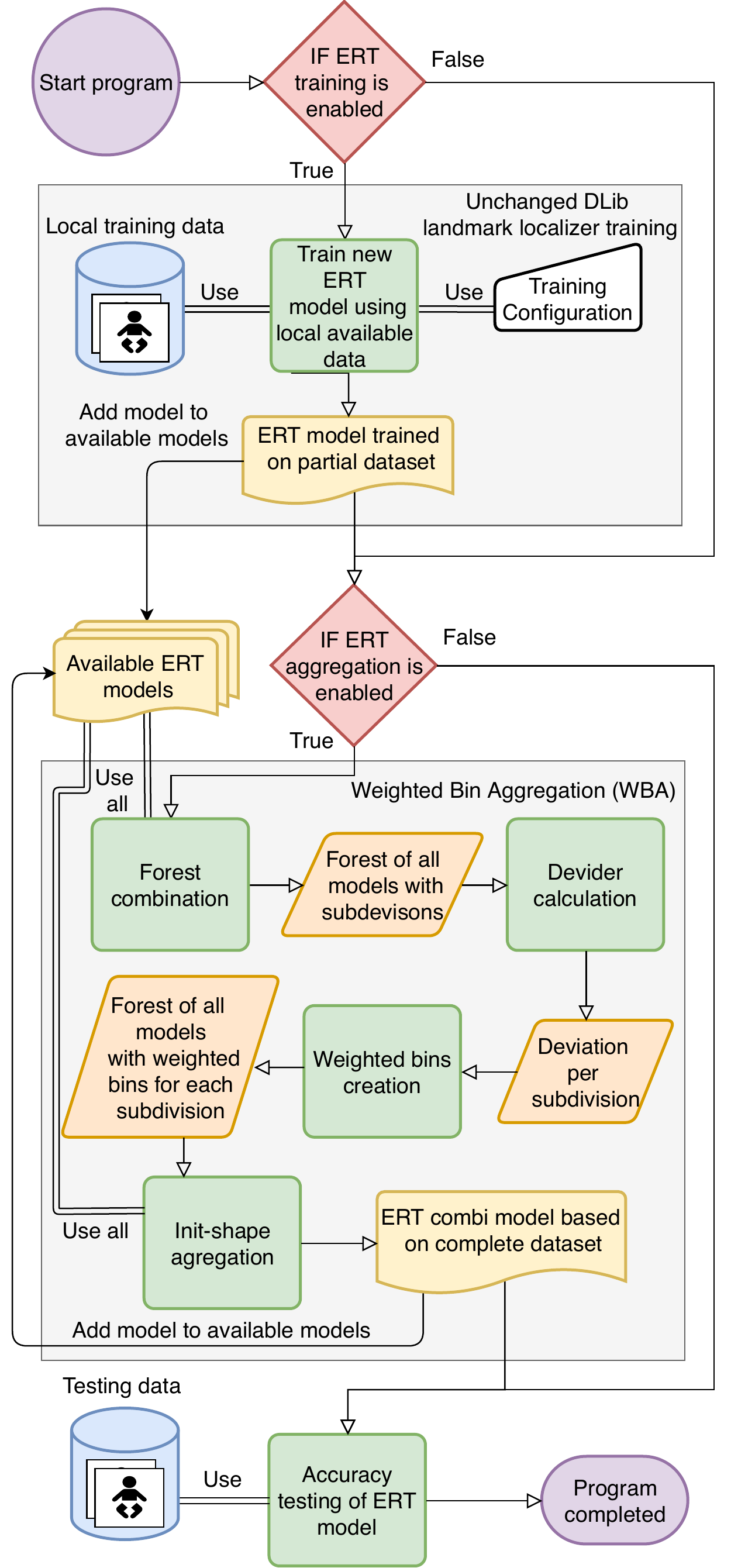}
    \caption{Landmark localizer model training and Weighted Bin Aggregation (WBA) overview for the ERT system. A new model can be trained from a selection of local training data, adding the available models together with models trained on other nodes. Available models are combined by WBA using forest aggregation to create a new model, where each forest represents a former model and has a weighted bin for its own sub-result. In the ERT process, the mean of these bins is used for the final result, therefore making use of all available training data without sending data between nodes.
 }
    \label{fig:dist_ert}
\end{figure}

The initial model of the ERT shape predictor and its trees are created during the training process. The currently used model is trained on the iBUG-300-W face landmark dataset~\cite{sagonas2013300, Sagonas2016} utilizing 68 landmarks. This dataset must be stored at a centralized server where the initial model is trained. 
In order to allow for making use of all training data without all data being shared, a Weighted Bin Aggregation (WBA) algorithm is proposed. The WBA algorithm is an extra layer on top of the training process and combines models derived from local datasets to form a new model which represents all training data. Where the models are combined depends on the  architecture for model distribution, which is covered in Section~\ref{sec:ditribute_arch}. The WBA-algorithm has a scalable performance when dividing model training across N nodes, with  a complexity of O(N), as each node can train its model in parallel and aggregation time is negligible compared to training. The WBA-algorithm and the training process is shown in Figure~\ref{fig:dist_ert}.

The training process itself is the same implementation of the DLib shape predictor training method. Models trained on local data from participating nodes are aggregated into a new model by combining the forests of regression trees and aggregating the shape used for initialization. This is achieved by making a new combined forest from the individual forests of the sub-models, where each forest of a sub-model forms its own subdivision represented by a weighted bin. The new initial shape is based on the average parts of each model. When the new model is used to localize landmarks, each subdivision calculates its own result regarding landmark position. Then a weighted mean is taken of these sub-results to form the final landmark coordinate. The aggregation process is visualized in Figure~\ref{fig:dist_ert} and described in more detail below. 

\begin{itemize}
\item \textbf{Forests combination}: All available forests from the models are collected and saved as one large forest with subdivisions, each representing the original used sub-model. For instance, if three models are used with 500 trees each, a new forest collection of 1500 trees is created with three subdivisions. 
\item \textbf{Divider calculation}: Each subdivision takes part in the end result. This can be symmetrical, where each bin contributes evenly or asymmetrically, where some bins contribute more than others, depending on configuration or sub-model test results. This allows for sub-models that are more accurate (because, for example, they were trained on a larger dataset) to have more weight in the end result than other sub-models. To indicate how much weight a subdivision has, a divider for the new model is calculated. The divider consists of the deviation for each subdivision and the total deviation value, which is needed to form the final landmark result. Following the example of three models used in the previous bullet, if the first model needs to have more weight than the other models, then the divider can be calculated as 1.5 deviation for model 1, and 1.0 for models 2 and 3, resulting with a final total deviation of 3.5.
\item \textbf{Weighted bins creation}: For each subdivision, a weighted bin is created. Each bin consists of a storage for the x and y values, and its deviation is given by the divider. If the new model is used, each subdivision calculates its own sub-results in parallel (according to the ERT steps described earlier), multiplies it by its own deviation, and stores it in the bin. To provide the end result, a weighted mean is calculated for both x and y values by dividing the summation of all bins by the total deviation value. This results in a coordinate representing a landmark, and this is repeated for all landmarks.
\item \textbf{Init-shape aggregation}: Each subdivision of the new model uses the same shape to initialize the ERT process. The shape used by the new model is an average from all used sub-models. For each part, a mean is used from the same part across all used sub-models.
\end{itemize}

In order to test the WBA algorithm, new models were created using the described distributed approach and compared to models resulting from conventional training. Training requires the use of a dataset with landmark and face annotation in XML formatting compatible with the DLib training process. The iBUG-300-W facial annotations dataset is selected. 
This consists of a training and testing dataset, where the training set is composed of 6666 images and the testing dataset of 1008 images. The training set is split into six equal parts, each consisting of 1111 images.  
We trained models for each part of training, one model for all six parts combined, and one for each iterative combination of parts until all are combined. For training, the parameters in Table~\ref{tab:ert_config} were used. The model based on all parts combined represents a model trained in a centralized server if all data are available. All sub-models were then combined using the aggregation approach described above in order to simulate distributed learning. 

In order to check for the impact of the order in which we aggregate sub-models, multiple aggregation orders are used. Accuracy is then tested for each model using the test dataset. As the ERT system moves its landmarks towards the real location, instead of either placing or not placing a landmark, accuracy is not measured in true or false positives. Instead accuracy is measured in how close the final landmarks are to their real counterparts using the mean error rate. The mean error rate is derived from the mean difference in distance between placed landmarks by the predictor and those of the annotated dataset:\\ 
\indent Mean error $={\frac {1}{n}}\sum _{i=1}^{n}($PLM$_{i}-$RLM$_{i})$,\\ 
\indent where PLM represents the placed landmark, RLM the real landmark according to data annotation, and $n$ the number of landmarks used by the model, in our case 68.

\begin{table}\centering
  \ra{1.3}
        \caption{Training settings used for distributed ERT testing 
        } \label{tab:ert_config}
\begin{tabular}{@{}lclc@{}}
    \toprule 
 \textbf{Setting} & \textbf{Parameter}\\
      \midrule 
    Oversampling            & 20 \\ 
    Nu                      & 0.1 \\ 
    Tree depth              & 5 \\ 
    Feature pool size       & 400 \\ 
    Number of test splits   & 20 \\ 
    Number of cascades      & 10 \\ 
    Trees per cascade       & 500 \\ 
    Lambda                  & 0.1 \\ 
    \end{tabular}
\end{table}

\subsection{Evaluation results}

Table~\ref{tab:ert_res} lists the accuracy results measured as the mean error rate for models trained on a partial dataset, models trained on the complete dataset (Pall), those on the steps in between, and those derived from WBA model combination (COM). The results show that models derived from the WBA have a lower mean error rate than models trained on a smaller dataset, e.g., the WBA combination of all models shows an 8\% improvement over the best trained local model, P3. However, the aggregated models show an increased mean error rate when compared to a model trained directly on the same centralized data, e.g., The WBA combination of all models has an increase of 17\% in mean error rate over Pall. It is estimated that the aggregated models are less assertive per level of the cascade when compared to direct training, due to smaller shifts in the landmarks. This means traditional learning has improved results and should be applied when possible, while WBA could be used when the data cannot be shared freely. In addition, the results show that WBA is independent of model input order. 

\begin{table}\centering
  \ra{1.3}
        \caption{Accuracy results for the traditional ERT and the WBA distributed training algorithm 
        } \label{tab:ert_res}
\begin{tabular}{@{}lclc@{}}
    \toprule 
     \textbf{ Dataset} & \textbf{Mean error rate} \\
      \midrule 
    iBUG-300-W-P1               & 0.0727 \\ 
    iBUG-300-W-P2               & 0.0730 \\ 
    iBUG-300-W-P3               & 0.0717 \\ 
    iBUG-300-W-P4               & 0.0722 \\ 
    iBUG-300-W-P5               & 0.0723 \\ 
    iBUG-300-W-P6               & 0.0728 \\ 
    iBUG-300-W-P1+P2            & 0.0656 \\ 
    iBUG-300-W-P1+P2+P3         & 0.0628 \\ 
    iBUG-300-W-P1+P2+P3+P4      & 0.0589      \\ 
    iBUG-300-W-P1+P2+P3+P4+P5   & 0.0572     \\ 
    iBUG-300-W-Pall             & 0.0568 \\ 
    iBUG-300-W-COM(P1+P2)       & 0.0690 \\ 
    iBUG-300-W-COM(P1+P2+P3)    & 0.0679 \\ 
    iBUG-300-W-COM(P1+P2+P3+P4) & 0.0665 \\ 
    iBUG-300-W-COM(P1+P2+P3+P4+P5)  & 0.0662 \\ 
    iBUG-300-W-COM(P1+P2+P3+P4+P5+P6)    & 0.0662 \\
    iBUG-300-W-COM(P6+P2+P4+P1+P5+P3)    & 0.0662 \\ 
    \end{tabular}
\end{table}

\section{Distribution architecture} \label{sec:ditribute_arch}
Both the proposed MWMA and WBA algorithms aggregate locally trained models, representing their local data with new models, which express the combined data. The aggregation processes of both algorithms are independent of location and model-distribution architecture, as both algorithms use models trained locally without dependency on results from other nodes. Moreover, the aggregation process does not have to take place on a certain node or to use the locally trained models in a certain order. Most related approaches in previous work can also be made location-independent, but some are more suitable for central processing, for example when the goal is to collect and process encrypted training data~\cite{yang2017privacy, wang2016sechog}. Independence means that the aggregation can take place at either node, or at the central server with or without own training data. This makes several configurations of the algorithms possible, by changing the model-distribution architecture, depending on operator choice or needed test-case functionality. For example, using a similar distribution architecture as in federated learning, a central-aggregation approach can be used, where locally trained models are sent to a central server, where aggregation takes place before sending the new model back to all nodes. A decentralized architecture is also possible, where each node is responsible for both training sub-models using locally available data, as well as aggregating using sub-models received from other nodes, leaving the choice of aggregation in the control of the node itself. 

\begin{figure}
    \centering
    \includegraphics[width=\linewidth]{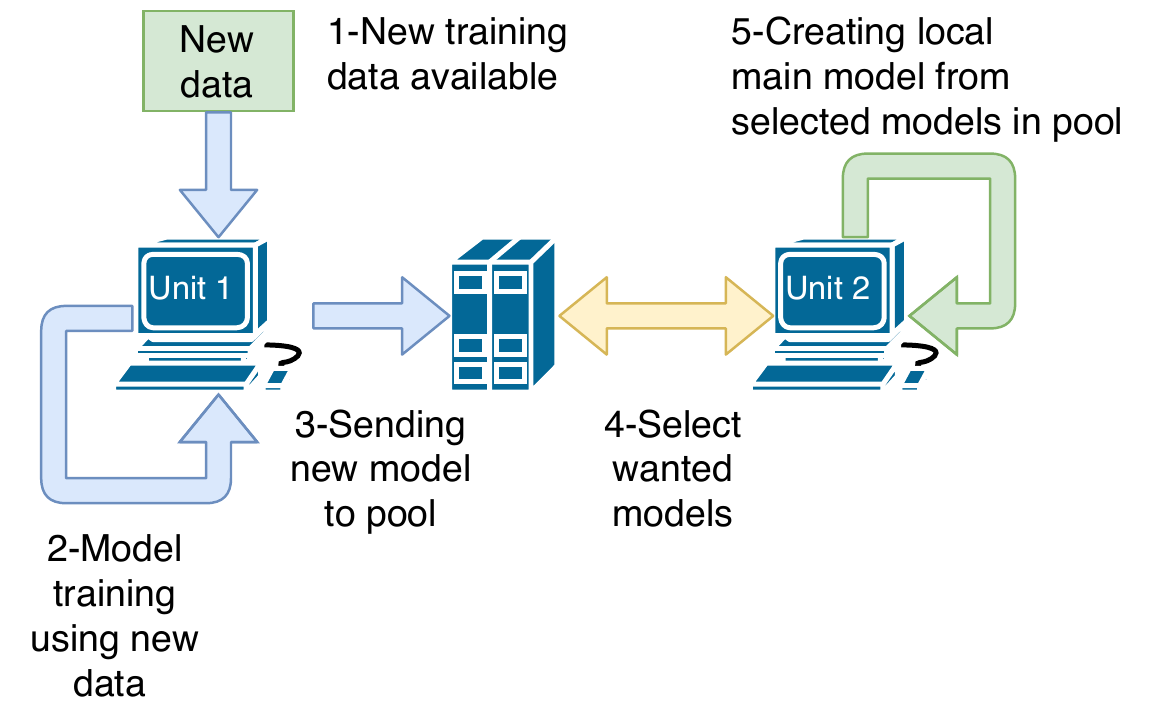}
    \caption{Overview of the Pool-Based Local Training and Aggregation (PBLTA) architecture for model distribution. Models are trained and combined locally. Created models are sent to a central pool server, allowing download from other nodes.}
    \label{fig:pool_dist}
\end{figure}

In this work, the Pool-Based Local Training and Aggregation (PBLTA) architecture for model distribution, Figure \ref{fig:pool_dist}, is proposed. The PBLTA is a hybrid solution between the central and decentralized approaches, where locally trained sub-models are sent to a central pool storage labeled with information regarding type of datasets used and achieved accuracy. Other nodes download these sub-models and aggregate them locally into a new model. The new models resulting from this aggregation can be uploaded to the pool as well, to share them with other nodes. Which sub-models are downloaded for use or aggregation by a node is completely dependent on operator choice, allowing total control over when a model is created and on what sub-models are used.

Other nodes are also not required to share sub-models with the pool, or to participate evenly; one node can share sub-models trained from 10000 images while another node can share a single sub-model trained on 600 images. The choice of what model to use for aggregation also allows weighing of the new model, biasing it toward a dataset of choice by including more models trained on that dataset or including a copy of those models in the aggregation process. As each node is responsible for the creation or downloading of the new main model, other nodes or the host of the pool are not required to preform maintenance work for other nodes. Each participant can focus independently on their own model.

The central pool itself is hosted by either one of the nodes or an untrusted third party. The pool does not have access to the training data, and only receives trained models with an information label. If a third party or rogue participant misuses access to the pool, models can leak to other parties. Leaking of trained models for both the HOG-algorithm and ERT is not a privacy concern, as training data cannot be extracted from that model. 

In the EBC case study, the PBLTA model-distribution architecture is used together with the MWMA and WBA algorithms to allow distributed training of models for the HOG-algorithm and ERT system, respectively. However, the PBLTA architecture does not require these specific algorithms for model aggregation and can use most other algorithms for local or central model aggregation. In a central-aggregation setting, models of choice are combined before being sent to the node, making the server responsible for the computational load needed for aggregation.

\section{Conclusions} \label{sec:conc}

This paper investigated the implementation of a distributed machine learning approach to train  models for both face detection and landmark localization from distributed training data, without sharing data between the collaborating research sites. The new algorithms were explored using an eyeblink conditioning (EBC) case study, for which new models needed to be trained. As training data for the EBC case study were distributed across both the Princeton Neuroscience Institute and the Erasmus Medical Center, and sharing of data between participants was not possible due to privacy concerns, traditional machine learning did not allow the full use of the training data. Distributed machine learning, on the other hand, does not require raw data to be sent between nodes.

This work proposed two new algorithms for distributing the training of models for the HOG-algorithm and an ERT landmark localizer, and showed how new models can be aggregated from models trained on local datasets to achieve distributed machine learning. Both proposed algorithms are independent of the location of their model aggregation and have linear scalable performance, allowing custom model distribution architectures. Both the WBA and MWMA algorithms and local model training software have been made available for public use on the following GitHub repository: https://github.com/SLWZwaard/DMT.

Results showed that for the HOG-algorithm it was possible to combine trained models using the proposed Mean Weight Matrix Aggregation (MWMA) algorithm, leading to an increase of recall accuracy of 0.9\% when compared to traditional training. The distribution of ERT training using the Weighted Bin Aggregation (WBA) algorithm was shown to be possible, while reducing the error rate by at least 8\% over the best individual locally trained model without data sharing, but with a 17\% increase in error rate compared to traditional training. 

For model distribution, a Pool-Based Local Training and Aggregation (PBLTA) architecture was proposed. Using pool-based storage, models can be shared between collaborators at different research sites for both algorithms. Models trained or aggregated locally can be shared to the pool, and downloaded models can be aggregated into new models. This allows collaborators to choose what models they want to use and what models they want to share with others. With new distributed algorithms to train, aggregate models, and the PBLTA architecture, new models can be created for the eyeblink conditioning paradigm without privacy concerns. While these algorithms were explored for this case study, they are not limited to facial landmark localization and can generalize to others forms of object detection and landmark localization. 


\ifCLASSOPTIONcaptionsoff
  \newpage
\fi



\bibliographystyle{IEEEtran}
\bibliography{IEEEabrv,references}
\end{document}